\documentclass[conference]{IEEEtran}
\IEEEoverridecommandlockouts
\usepackage{cite}
\usepackage{amsmath,amssymb,amsfonts}
\usepackage{algorithmic}
\usepackage{graphicx}
\usepackage{textcomp}
\usepackage{multirow}
\usepackage{multicol}
\usepackage{setspace}
\usepackage{diagbox}
\usepackage{enumitem}
\usepackage{float}
\usepackage{subfigure}
\def\BibTeX{{\rm B\kern-.05em{\sc i\kern-.025em b}\kern-.08em
    T\kern-.1667em\lower.7ex\hbox{E}\kern-.125emX}}

\begin{document}

\title{Mirror-Yolo: A Novel Attention Focus, Instance Segmentation and Mirror Detection Model}

\author{\IEEEauthorblockN{Fengze Li$^{1,2,3}$, Jieming Ma$^{2*}$, Zhongbei Tian$^1$, Ji Ge$^3$, Hai-Ning Liang$^2$, Yungang Zhang$^4$, Tianxi Wen$^2$}
\IEEEauthorblockA{
\textit{University of Liverpool, Liverpool, UK$^1$}\\
\textit{Xi'an Jiaotong-Liverpool University, Suzhou, China$^2$}\\
\textit{JITRI Micro-Nano Automation Institute, Suzhou, China$^3$}\\
\textit{Yunnan Normal University, Kunmimg, Yunnan, China$^4$}\\
$^*$Corresponding author's email: Jieming.Ma@xjtlu.edu.cn}
}

\maketitle

\begin{abstract}
Mirrors can degrade the performance of computer vision models, but research into detecting them is in the preliminary phase. YOLOv4 achieves phenomenal results in terms of object detection accuracy and speed, but it still fails in detecting mirrors. Thus, we propose Mirror-YOLO, which targets mirror detection, containing a novel attention focus mechanism for features acquisition, a hypercolumn-stairstep approach to better fusion the feature maps, and the mirror bounding polygons for instance segmentation. Compared to the existing mirror detection networks and YOLO series, our proposed network achieves superior performance in average accuracy on our proposed mirror dataset and another state-of-art mirror dataset, which demonstrates the validity and effectiveness of Mirror-YOLO.
\end{abstract}


\section{Introduction}
Mirrors are commonplace in day-to-day life, but very few studies have taken mirrors into account as a factor that can seriously reduce the performance of existing computer vision tasks. For years, many efficient object detection or segmentation models have been proposed along with the emergence of new approaches or tricks in one or several modules of the network structure. For instance, the prevalence of the classical R-CNN series \cite{girshick2014rich_rcnn,girshick2015_fastrcnn,he2017_maskrcnn}, YOLO series \cite{YOLO,YOLOv2,YOLOv3,YOLOv4} and Single Shot MultiBox Detector \cite{liu2016ssd} in the study of object detection. Additionally, there is some recent work contributing to field of obstacle detection, such as a new autoencoder-based approach to road obstacle detection by Toshiaki et al. in 2021 \cite{ohgushi2020roadobstacles}, and a YOLO-based network for obstacle detection in living environments that may be encountered by electric wheelchairs by Phenphitcha et al. in 2022 \cite{patthanajitsilp2022obstacles}. However, neither of them considered mirrors as obstacles for computer vision models, even though the reflected content of mirrors would make the detection much less accurate.

\begin{figure}[t]
\centering
\subfigure[wrong detection]{
\label{Fig.wrongdec}
\includegraphics[width=0.3\columnwidth]{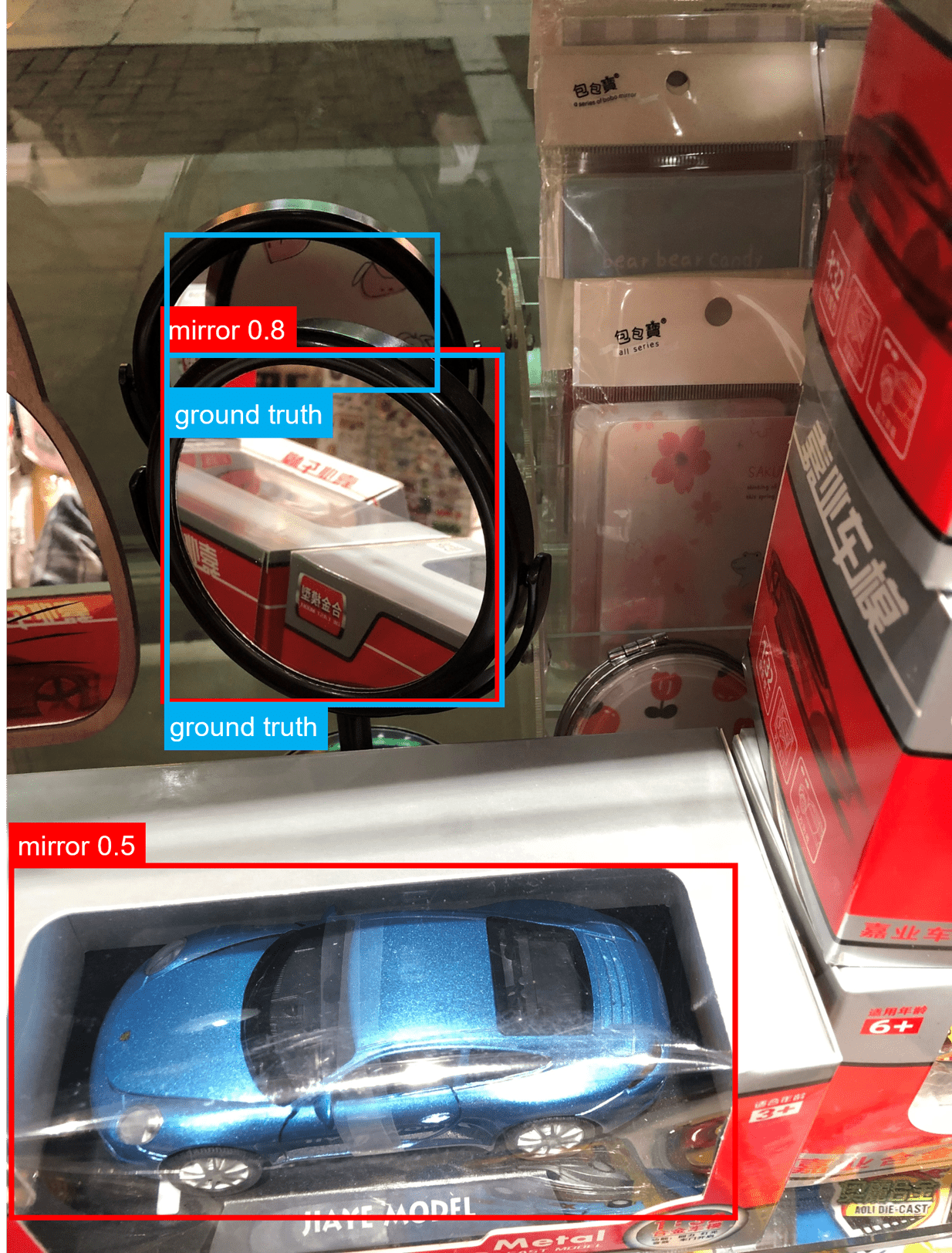}}
\subfigure[missed detection]{
\label{Fig.misseddec}
\includegraphics[width=0.3\columnwidth]{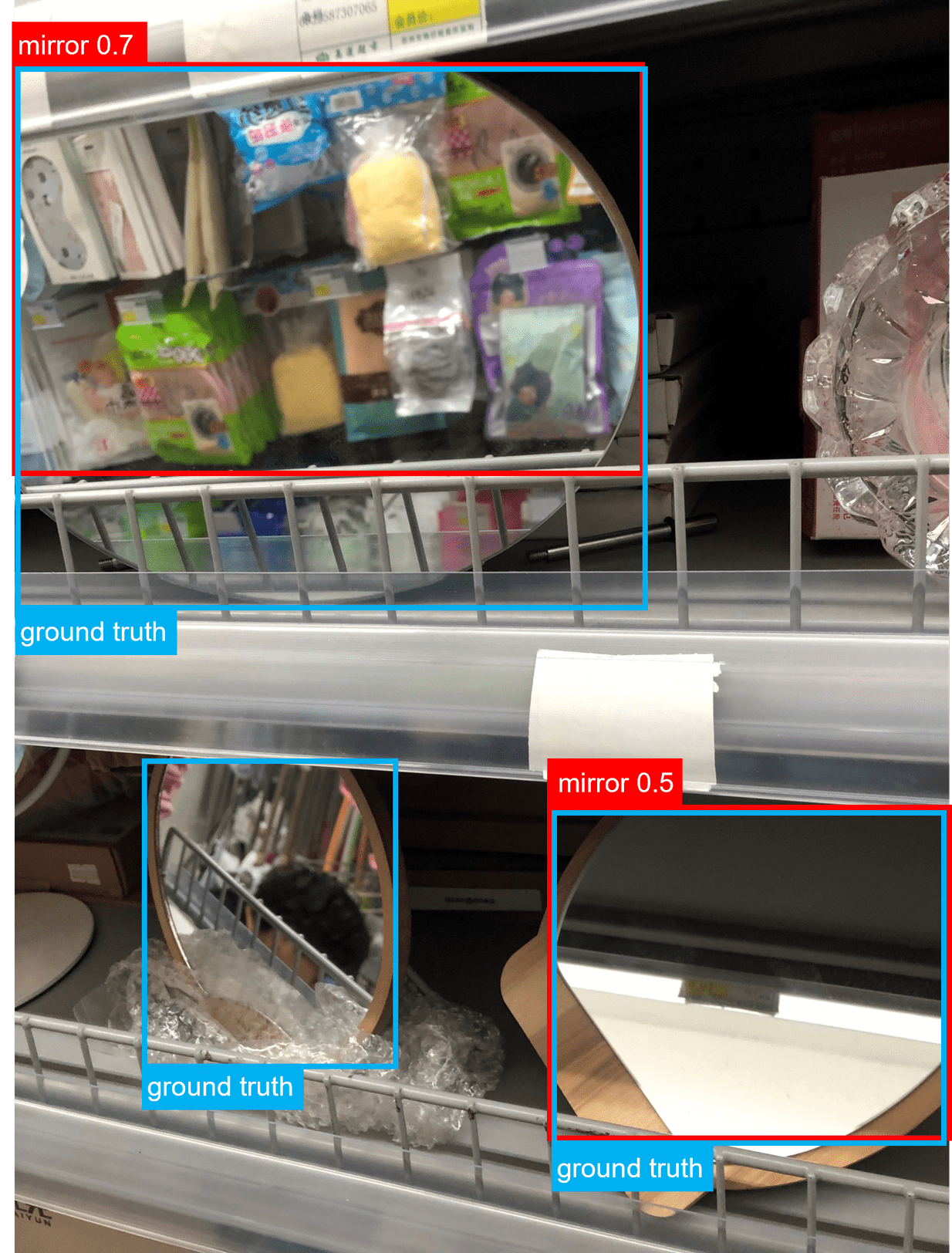}}
\subfigure[failed detection]{
\label{Fig.faileddec}
\includegraphics[width=0.3\columnwidth]{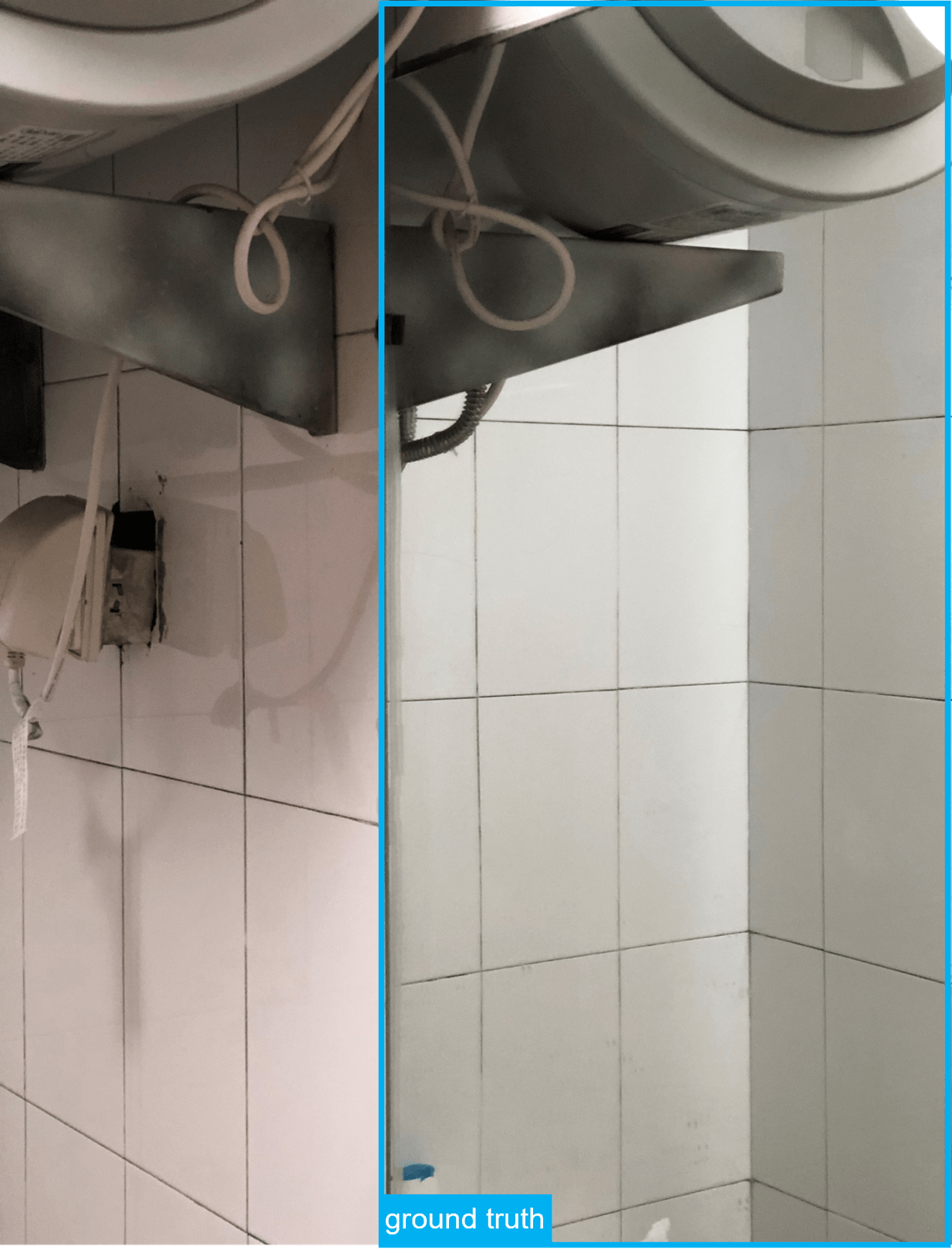}}
\caption{Problems of detecting mirrors by YOLOv4.}
\label{Fig.YOLO4problems}
\end{figure}

When only the mirror is of concern, Yang et al. first modeled mirror detection through low-level semantic features such as incoherence of color or texture \cite{yang2019my}, while Lin et al. instead focused on the distinction and integration of mirror and context features inside and outside the mirror \cite{lin2020progressive}. Yet, they all have limitations and scope for further enhancement. YOLOv4 \cite{YOLOv4} gets a significant performance improvement over the YOLO series to which it belongs, but it also failed to achieve satisfactory results on the task of mirror detection, shown in Fig. \ref{Fig.YOLO4problems}. 

\begin{figure*}[t]
    \centering
    \includegraphics[width=1\textwidth]{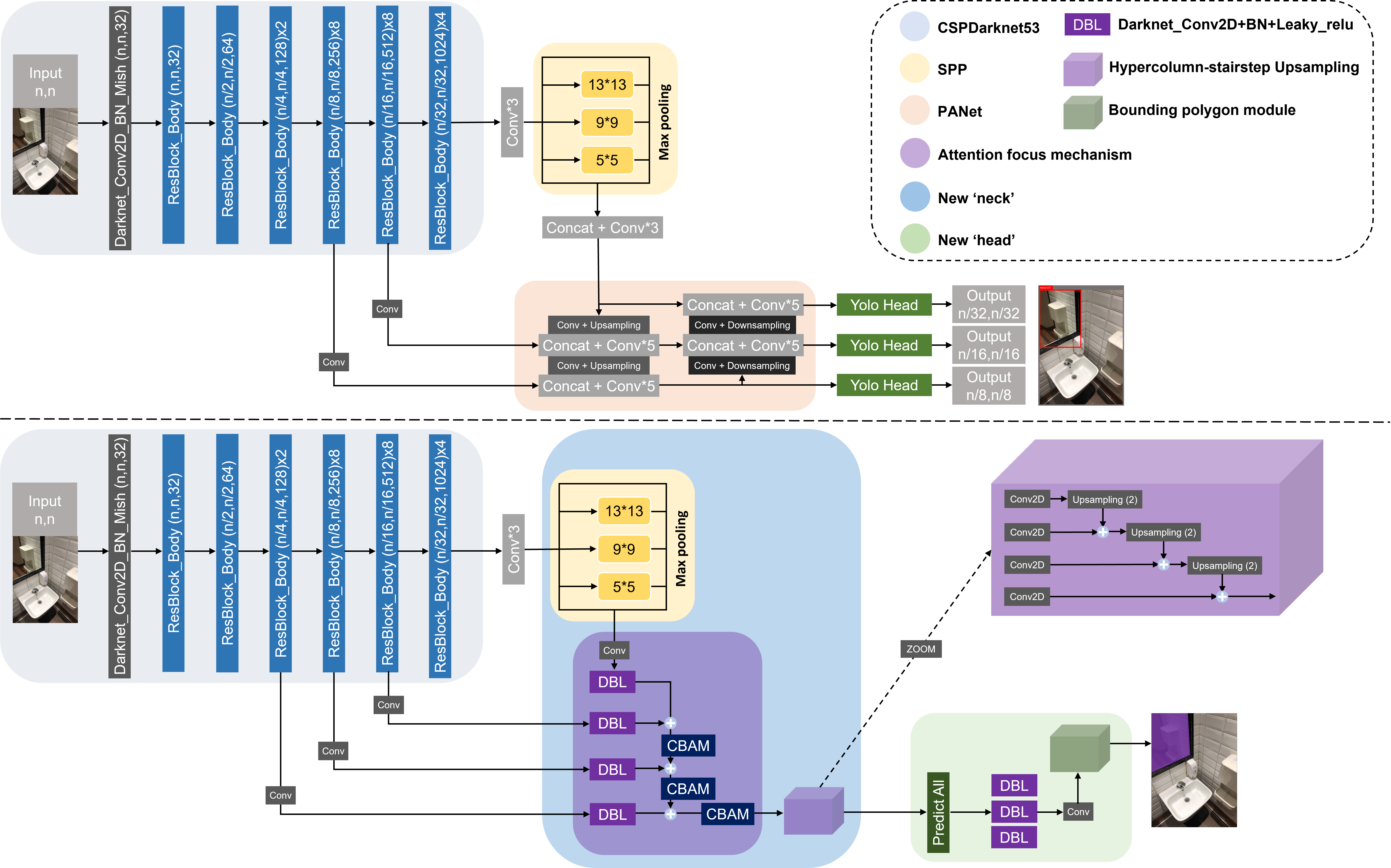}

        \caption{The network comparison of the origin YOLOv4 (top) and Mirror-YOLO (bottom).}
    \label{mirroryoloandv4}
\end{figure*}

One reason humans are able to detect mirrors accurately is that we can easily realise the discontinuity between what is in the mirror and its surroundings \cite{yang2019my}. Another more significant reason is that a wealth of experience enables people to concentrate on the parts of the image that matter and not be interfered with by other features that are potentially distracting. For neural networks, the importance of different features at different locations in the network (different network layers, or different channels, etc.) varies. With limited computational resources, the later layers should focus more on the critical information and suppress the unimportant ones. Hu et al. proposed Squeeze-and-Excitation (SE) in SENet \cite{hu2018squeeze_SENet} to select features by considering the weights of channels, while Convolutional Block Attention Module (CBAM) \cite{woo2018cbam} can combine spatial attention and channel attention. Also, Li et al., in thier recent study \cite{li2020object_attention_nature}, pointed out that either global average or max pooling is advantageous for channel-wise attention, and Lun et al. \cite{huang2019adaptively_aat} proposed Adaptive Attention Time (AAT), a novel adaptive attention model for image captioning via aligning the source and the target. What is more, using mirror bounding polygons for instance segmentation is illuminating. Poly-YOLO proposed by Hurtik et al. generates a flexible number of points that define an object's bounding polygon, which allows the network to be trained for various shapes, independent of the object's size \cite{hurtik2020poly_yolo}.

Therefore, we believe to embed an attention mechanism into YOLOv4, and with appropriate improvements on network architecture will make the original YOLOv4 more competitive for mirror detection tasks. The Mirror-YOLO (MY) proposed in this paper enhances the detection performance of specular objects in real-life environments. The main contributions of this work are listed as follows:
\begin{enumerate}
    \item Introducing an attention focus mechanism for better features acquisition and investigating the results brought about by different attention mechanisms in different network positions.
    \item Optimising the initial neck of the model by the hypercolumn-stairstep approach.
    \item Implementing the mirror bounding polygons module in the origin head to accomplish instance segmentation, reaching a broader range of application scenarios.
    \item Proposing a robust mirror dataset, which consists of 4,518 mirror images from multiple scenarios.
\end{enumerate}
% -----------------------------------------------------------------------------
\section{The Proposed Methods}
\label{sec:METHODS}
The overall architecture of our proposed Mirror-YOLO is shown in the lower part of Fig.\ref{mirroryoloandv4}. The brand new neck including an attention focus mechanism and hypercolumn-stairstep, and the head including the mirror bounding polygon module, compose the proposed Mirror-YOLO. All the components will be detailed subsequently.

\subsection{Attention Focus Mechanism Module}
\begin{figure}[t]
\centering
\subfigure[]{
\label{Fig.cbam1}
\includegraphics[width=0.45\columnwidth]{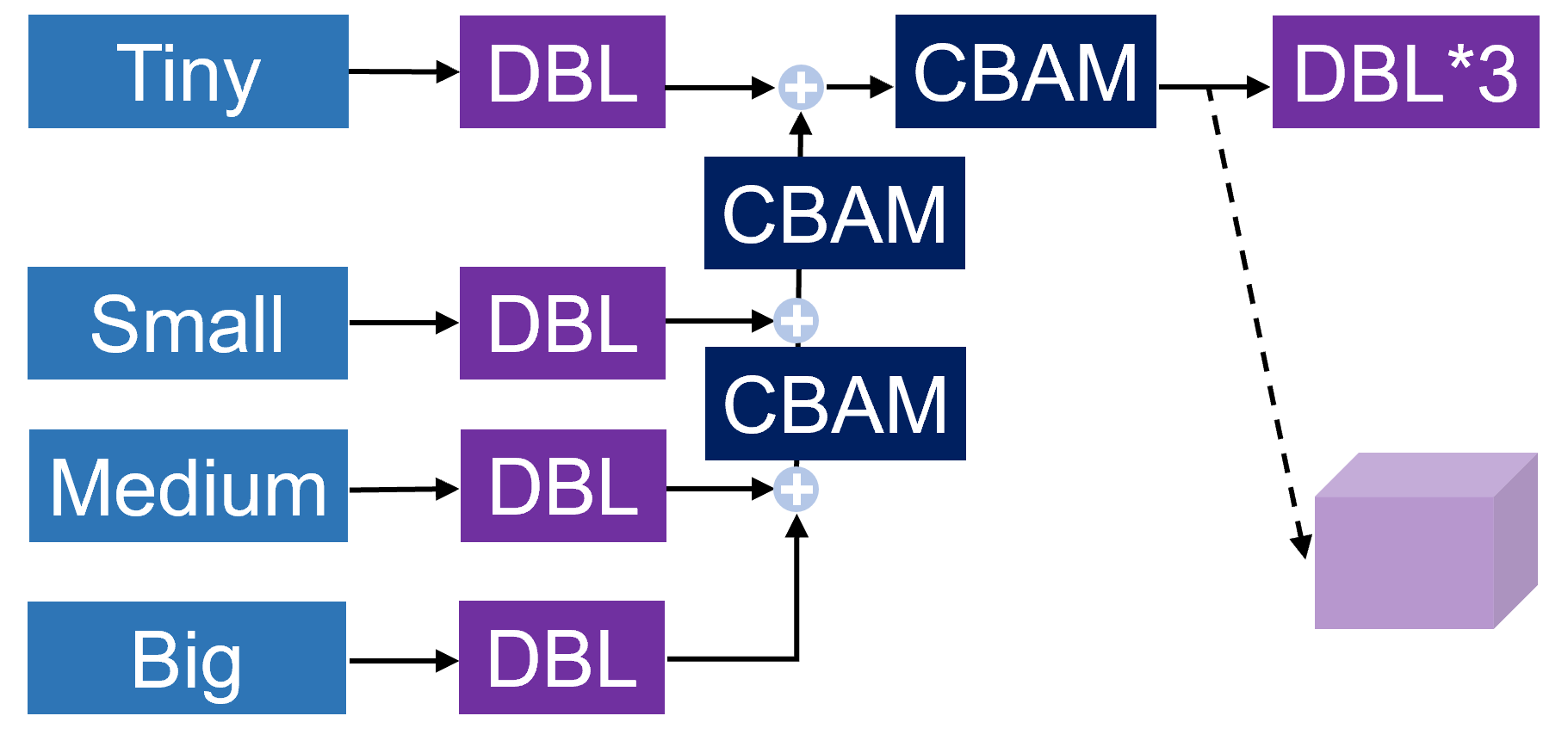}}
\subfigure[]{
\label{Fig.cbam2}
\includegraphics[width=0.45\columnwidth]{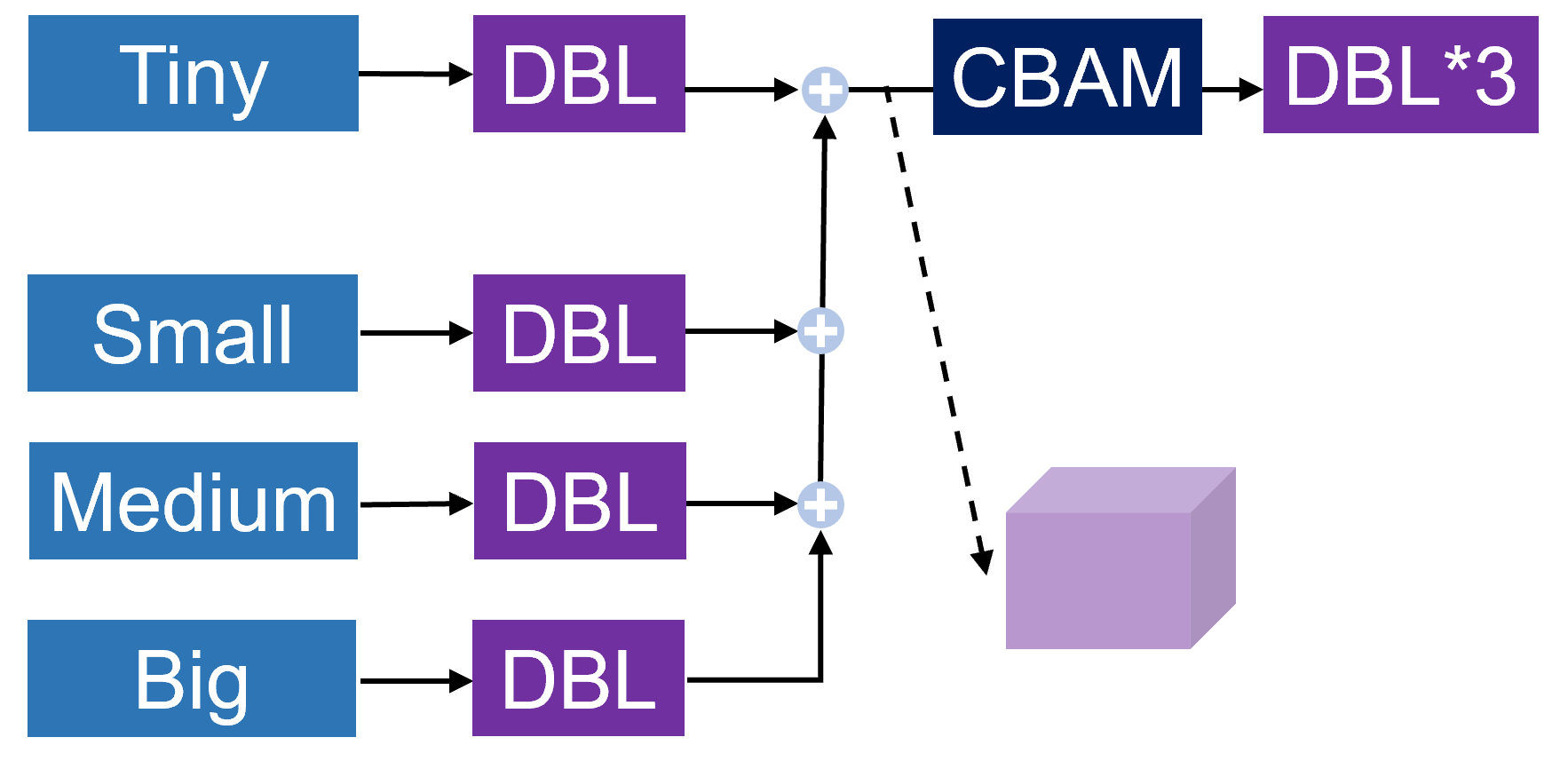}}\\
\subfigure[]{
\label{Fig.cbam3}
\includegraphics[width=0.45\columnwidth]{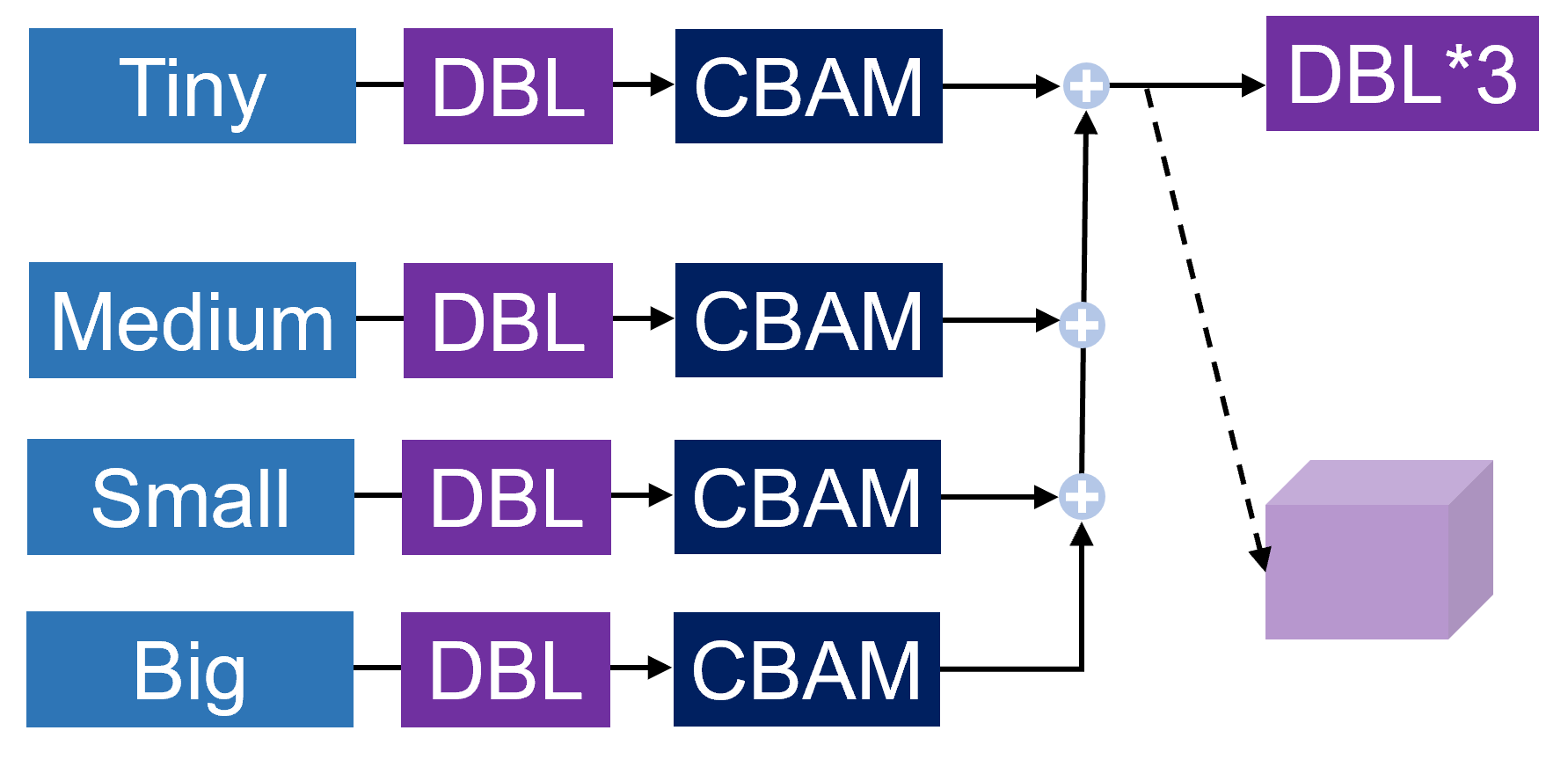}}
\subfigure[]{
\label{Fig.cbam4}
\includegraphics[width=0.45\columnwidth]{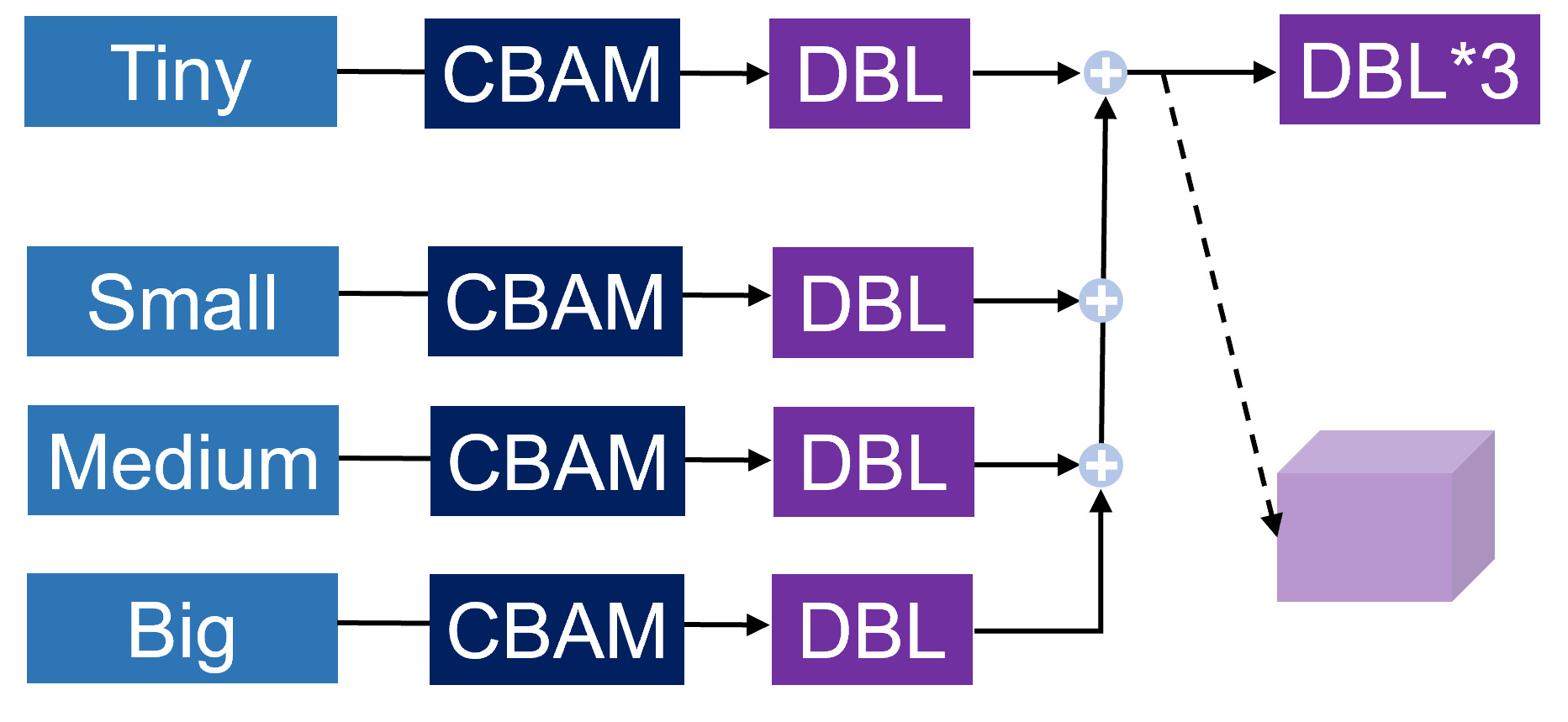}}
\subfigure[CBAM]{
\label{Fig.cbam_network}
\includegraphics[width=0.95\columnwidth]{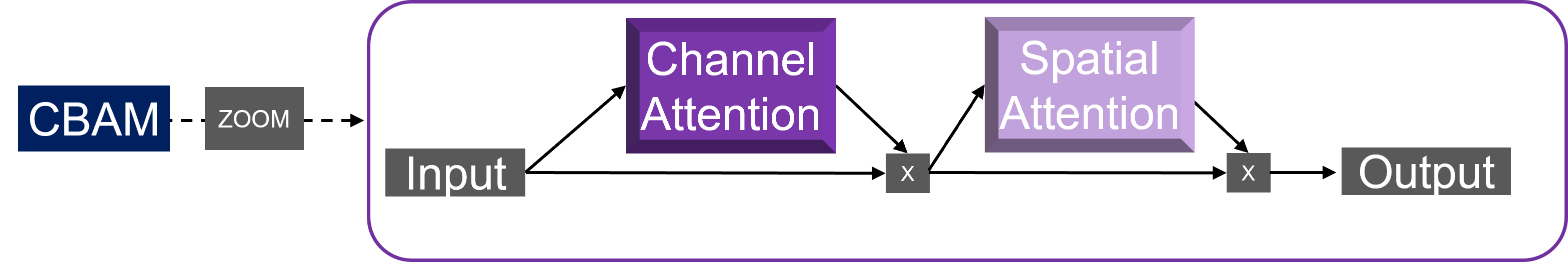}}\\
\subfigure[SE]{
\label{Fig.se_network}
\includegraphics[width=0.95\columnwidth]{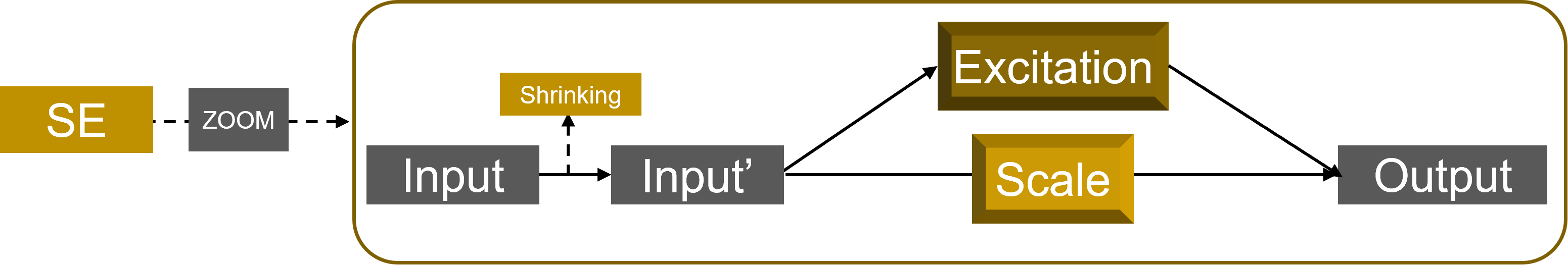}}
\caption{Our implementation (a) of CBAM module compared with its counterparts. The network structure of CBAM (e) and SE (f).}
\label{Fig.cbam_module}
\end{figure}

The introduced attention mechanism CBAM combines the attention mechanisms of both dimensions from channel feature and spatial feature, which can be described in the Fig.\ref{Fig.cbam_network}. We carry forward the demonstration of \cite{woo2018cbam} that the channel attention module in front of the spatial attention module leads to better experimental results. From the straightforward network structure of CBAM, the calculation method can be summarized as (\ref{equ_1}):
\begin{equation}
\begin{array}{r}
F^{\prime}=F \otimes M_{c}(F), \\
F^{\prime \prime}=F^{\prime} \otimes M_{s}\left(F^{\prime}\right),
\end{array}
\label{equ_1}
\end{equation}

where the input feature map is $F$,  the ultimate result of CBAM is $F^{\prime \prime}$, and $\otimes$ means element-wise multiplication. After element-wise multiplication of the input features with the weights generated by channel attention $M_{c}(F)$ , the channel weighting result represented by $F^{\prime}$ can be calculated, and correspondingly, the final result is so obtained after spatial attention. In terms of the modules within, equation (\ref{equ_2}) and (\ref{equ_3}) below demonstrate the computing method of channel and spatial attention module separately:
\begin{equation}
\begin{aligned}
\mathbf{M}_{\mathbf{c}}(\mathbf{F}) &=\sigma(\operatorname{MLP}(\operatorname{AvgPool}(\mathbf{F}))+M L P(\operatorname{MaxPool}(\mathbf{F}))) \\
&=\sigma\left(\mathbf{W}_{\mathbf{1}}\left(\mathbf{W}_{\mathbf{0}}\left(\mathbf{F}_{\mathbf{a v g}}^{\mathbf{c}}\right)\right)+\mathbf{W}_{\mathbf{1}}\left(\mathbf{W}_{\mathbf{0}}\left(\mathbf{F}_{\mathbf{m a x}}^{\mathbf{c}}\right)\right)\right),
\end{aligned}
\label{equ_2}
\end{equation}
\begin{equation}
\begin{aligned}
\mathbf{M}_{\mathbf{S}}(\mathbf{F}) &=\sigma\left(f^{7 \times 7}([\operatorname{AvgPool}(\mathbf{F}) ; \operatorname{MaxPool}(\mathbf{F})])\right) \\
&=\sigma\left(f^{7 \times 7}\left(\left[\mathbf{F}_{\mathbf{a v g}}^{\mathbf{s}} ; \mathbf{F}_{\mathbf{m a x}}^{\mathbf{s}}\right]\right)\right),
\end{aligned}
\label{equ_3}
\end{equation}
where $\mathbf{F}$, $\mathbf{F}_{\mathbf{a v g}}^{\mathbf{c}}$, $\mathbf{F}_{\mathbf{m a x}}^{\mathbf{c}}$ represent feature map, feature after the computation of global average and global max pooling, respectively. $\mathbf{W}_{\mathbf{0}}$ and $\mathbf{W}_{\mathbf{1}}$ denote two layers of parameters in the multilayer perceptron, and $f^{n \times n}$ stands for a $n \times n$ convolution kernel.
% \begin{figure}[h]
%     \centering
%     \includegraphics[width=1\columnwidth]{fig/regular_our_CBAM.png}
    
%         \caption{Our implementation (right) of CBAM module and its counterparts.}   
%     \label{mirroryoloandv4}
% \end{figure}
% ---------------------------------------------------------------
This paper discussed the merits and demerits of the results obtained by the attention focus mechanism at four different locations in the proposed network structure, shown in Fig.\ref{Fig.cbam_module}, where Fig.\ref{Fig.cbam1} shows what we have proven to work optimally in the end, which is each feature layer goes through a DBL block, adding a method of the attention mechanism module after each fusion. In addition, this paper has also considered the potential change in effect if the CBAM is replaced with the SE module, which structure is shown in Fig.\ref{Fig.se_network}, and the subsequent ablation experiments (See Table \ref{Ablation}) will demonstrate the effectiveness of the CBAM.

% ------------------------------------------------------------------------------
\subsection{Hypercolumn-stairstep}
The original neck of YOLOv4 was added with the aforementioned attention mechanism and hypercolumn-stairstep module, where the hypercolumn-stairstep module takes the feature maps generated at the various levels of the backbone as input. Furthermore, such a structure differs from uniformly resizing the feature maps to the same resolution and adding them together, but instead resizing the lowest resolution feature map, adding it to the second lowest resolution feature map, and so forth then resizing the output resolution, shown in the lower part of Fig.\ref{mirroryoloandv4}. The output of this upsampling approach is smoother and more addings are realized. When $F$ represents the feature map of an input, it can be defined as:
% Furthermore, such a structure differs from uniformly resizing the feature maps to the same resolution and adding them together, but instead resizing the lowest resolution feature map, adding it to the second lowest resolution feature map. So forth, then resizing the output resolution, the procedure is shown in the lower part of Fig.\ref{mirroryoloandv4}. The outcome of this upsampling approach is more smoother, and more addings are realised. Let $F$ represents the feature map of an input, it can be defined as:
\begin{equation}
F^{\prime}=\ldots u\left(u\left(m\left(F_{n}\right), 2\right)+m\left(F_{n-1}\right), 2\right) \cdots+m\left(F_{1}\right),
\label{equ_4}
\end{equation}
where $m$ is a function used to convert the dimensions of a feature map from $a \times b \times c \times -$ to $a \times b \times c \times \delta$ with constant $\delta$, and $u$ denotes a function that upscales the input. Such a procedure is where the model obtains a lower loss without increasing the computational complexity and time consumption.

% ------------------------------------------------------------------------------
\subsection{Mirror Bounding Polygons}
\begin{figure}[h]
    \centering
    \includegraphics[width=1\columnwidth]{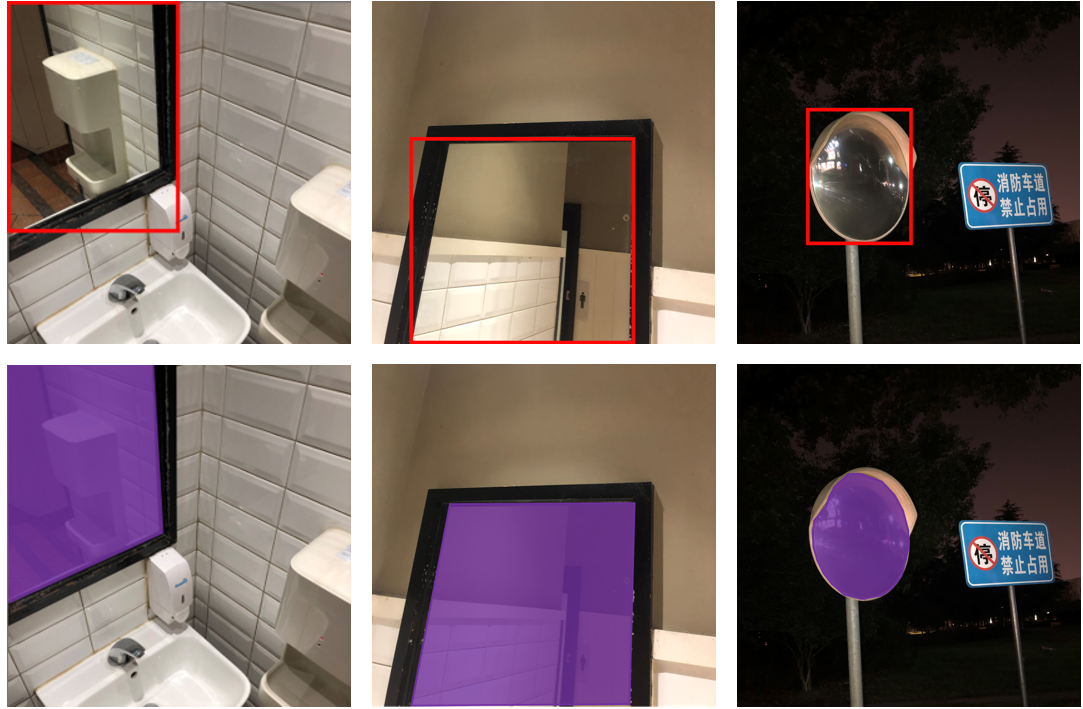}
        \caption{Comparison between the results obtained by bounding boxes (top) and our mirror bounding polygons (bottom).}
    \label{polygons}
\end{figure}

Unlike YOLOv4, where the predicted anchors are obtained by raster mask, the embedding of the mirror bounding polygons enables Mirror-YOLO to produce a flexible set of points that define the bounding of an object, as shown in Fig. \ref{polygons}.

In detail, the vertices are detected as when an object and its bounding box are detected. Each vertex is given its own confidence, which is compared to a threshold value, and ultimately, those vertices whose confidence is above a threshold define the bounding polygon. As with the bounding box coordinates of the YOLO series, the vertices are generated by the output layer, which means the remaining parts of the network architecture are not affected by this module, so the number of parameters added by the bounding polygon is negligible. Notably, any dataset with semantic segmentation information can generate labels for the described vertices. Thus, this method of enriching each object label with a series of vertices of bounding polygons enables the network to be trained for general ones, independent of the object's shape.
% ------------------------------------------------------------------------------
\begin{table*}[t]
	\centering
	\caption{The comparison between Mirror-YOLO and other methods based on our dataset and MSD.}
	\setlength\tabcolsep{7pt}
	\renewcommand{\arraystretch}{1.1}
	\begin{tabular}{c|cccc|cccc}
		\hline
		\multirow{2}{*}{Algorithms}&\multicolumn{4}{c|}{Our Dataset}&\multicolumn{4}{c}{Original MSD Dataset}\\\cline{2-9}
 & {MAE$\downarrow$} & {F$_\beta\uparrow$} & E-Measure$\uparrow$ & S-Measure$\uparrow$ & {MAE$\downarrow$} & {F$_\beta\uparrow$} & E-Measure$\uparrow$ & S-Measure$\uparrow$\\ \hline
        PSPNet\cite{8100143psp}   &   0.138     &    0.715  & 0.771 &  0.750 & 0.138 & 0.717 & 0.767 & 0.752   \\\hline
        PoolNet\cite{Liu2019PoolSal_pool}   &  0.146  &    0.679  &  0.708  & 0.695 & 0.151 & 0.655 & 0.682 & 0.671  \\\hline
        Mask-RCNN\cite{he2017_maskrcnn} &   0.149     &  0.668  &  0.692 & 0.693 & 0.143 & 0.672 & 0.699 & 0.697    \\\hline
        YOLO3\cite{YOLOv3}   &   0.109     &    0.763& 0.851 & 0.904 & 0.112 & 0.762 & 0.844 & 0.895    \\\hline
        YOLO4\cite{YOLOv4}   &   0.112     &    0.770  &   0.854 & 0.908 & 0.110 &0.771 & 0.854 & 0.910  \\\hline
        \textbf{Mirror-YOLO} &   \textbf{0.087} &    \textbf{0.822}  &  \textbf{0.873} &  \textbf{0.926} & 0.084 & 0.831 & 0.877 & 0.934  \\\hline
	\end{tabular}
	\label{final_results}
\end{table*}
\section{Experiments and Results}
\label{sec:EXPERIMENTAL RESULTS}
Our experiments are implemented by using the Tensorflow framework, and run on an NVIDIA Tesla T4 from Google Colab. To validate the performance of the proposed Mirror-YOLO, experiments were set up to compare representative YOLO models with Mirror-YOLO, the compared methods were trained and tested in the same settings. Moreover, an ablation experiment was implemented on the proposed attention focus module to verify the validity of the proposed method.

To the best of our knowledge, there is only one open source and available mirror dataset, MSD \cite{yang2019my}. It covers 4,018 images with mirrors along with the corresponding masks. However, the MSD mainly contains indoor scenes, and the dataset has a relatively high similarity between images. Thus, we constructed a new dataset based on the MSD, with 500 new mirror images containing both indoor and outdoor scenes added, most of which were taken by mobile devices. A small portion of images is taken from the ADE20K \cite{zhou2019semantic_ade20k} bathroom dataset.
\begin{table}[h]
	\centering
	\caption{The comparison of similarity score between MSD and ours.}
	\begin{tabular}{c|cr}
		\hline
		Dataset    &  MSD & Ours  \\ \hline

        Similarity score  &   34.7\%      &    \textbf{31.2\%}      \\\hline
	\end{tabular}
	\label{Simi}
\end{table}

The similarity between the mirror images from the MSD dataset and our dataset is compared in Table \ref{Simi}. All images were adjusted to a same scale and the image similarity is measured by SSIM \cite{wang2004imageSSIM}. The ratio of the sum of the SSIMs of a pair of images from the dataset to the number of pairs is the similarity score we presented, shown as below:
\begin{equation}
\text { Similarity score }=\frac{\sum_{n=1}^{N} S S I M\left(x_{n}, y_{n}\right)}{N},
\end{equation}
where SSIM is calculated with reference to the source paper \cite{wang2004imageSSIM} for details. $N$ denotes the total number of image pairs and $n$ represents the $n^{th}$ pair. As shown in Table \ref{Simi}, our dataset achieves a lower similarity.

\subsection{Ablation Experiment}
In order to compare the effect of different attention mechanisms at different locations (Fig. \ref{Fig.cbam_module}), an ablation experiment was implemented, the results are shown in Table \ref{Ablation}.
\begin{table}[b]
	\centering
	\caption{Ablation experiments on the effect of different network locations of different attention mechanisms on results.}
	\setlength\tabcolsep{3pt}
	\renewcommand{\arraystretch}{0.85}
	\begin{tabular}{c|c|c|c}\hline
        \diagbox[height=16pt]{F$_\beta\uparrow$}{MAE$\downarrow$} & MY & MY + CBAM & MY + SE \\\hline
        -   &  \diagbox[height=16pt]{0.771}{0.112}  & -  & - \\\hline
        Location (a)   &    -    &    \diagbox[height=16pt]{\textbf{0.822}}{\textbf{0.087}}   & \diagbox[height=16pt]{0.814}{0.093}  \\\hline
        Location (b)   &   -    &    \diagbox[height=16pt]{0.806}{0.997}   & \diagbox[height=16pt]{0.797}{0.998}  \\\hline
        Location (c)   &    -  &    \diagbox[height=16pt]{0.765}{0.102}  & \diagbox[height=16pt]{0.790}{0.109}  \\\hline
        Location (d)   &    -     &    \diagbox[height=16pt]{0.777}{0.113}  & \diagbox[height=16pt]{0.781}{0.105} \\\hline
	\end{tabular}
	\label{Ablation}
\end{table}

As can be seen in Table \ref{Ablation}, the introduction of the attention mechanism apparently bring positive effect for the detection performance. Although the SE and the CBAM made no significant difference in mirror detection tasks, it was demonstrated in our experiment that the embedding of CBAM as in Fig. \ref{Fig.cbam1} is the most beneficial option for the proposed model.

\subsection{Experimental Results of Different Detection Models}
Six object detection models were compared for mirror detection on our mirror image dataset. The results were evaluated by four different metrics: MAE, F$_\beta$, E-Measure \cite{ijcai2018-97_emeasure} and S-Measure \cite{Fan_2017_ICCV_smeasure}, the results are shown in Table \ref{final_results}.

The experimental results show that our proposed Mirror-YOLO outperforms the standard YOLOv4 and other object detection networks in each evaluation metric. Furthermore, the results show that some state-of-the-art salient object detection models such as the PSPNet\cite{8100143psp} and the PoolNet\cite{Liu2019PoolSal_pool} still have limitations in mirror detection. Compared to the original YOLOv4, the proposed Mirror-YOLO improves the F$_\beta$ metric by 6.7\% and the MAE metric by 21.7\% in mirror detection.

In addition, experiments conducted with the same network on different datasets have shown that the robustness of the dataset facilitates more reliable results. Even though there are minor differences in the experimental results on the two datasets, we consider them to be negligible. Overall, our proposed model presents outstanding results on both datasets, thus the validity and sophistication of the model is proven.

\section{Conclusion}
In this paper, Mirror-YOLO based on the YOLOv4 backbone was proposed to address the problem of mirror detection. It fuses a structurally optimized attention mechanism, hypercolumn-stairstep, and mirror bounding polygons to achieve high accuracy in mirror detection and instance segmentation. On this basis, this article explores the effects of various attentional mechanisms in different network locations on the experimental results through ablation experiments. The paper also proposed a novel mirror dataset with a low repetition rate and high robustness. Comparative experiments have demonstrated the validity of Mirror-YOLO compared to other state-of-the-art object detection networks. 

The experimental results were obtained separately on two different datasets. The success on multiple datasets confirms the superiority of Mirror-YOLO in the field of mirror detection. More importantly, this method can also be used as one of the pre-processing steps for industrial computer vision tasks including autonomous driving and robot path planning, while also offering algorithmic support and theoretical validation for subsequent mirror detection for or reconstruction in the 3D community.

\section*{ACKNOWLEDGEMENTS}
This research is supported by the Suzhou Science and Technology Project-Key Industrial Technology Innovation (Grant No. SYG202006, SYG202122), Future Network Scientific Research Fund Project (FNSRFP-2021-YB-41), the Key Program Special Fund of Xi’an Jiaotong-Liverpool University (XJTLU), Suzhou, China (Grant No. KSF-E-65), and the XJTLU AI University Research Centre and Jiangsu (Provincial) Data Science and Cognitive Computational Engineering Research Centre at XJTLU.

\bibliographystyle{IEEEbib}
\bibliography{refs}

\end{document}